\documentclass[10pt,twocolumn,letterpaper]{article}

\usepackage{cvpr}              

\usepackage{graphicx}
\usepackage{amsmath}
\usepackage{amssymb}
\usepackage{booktabs}
\usepackage{multirow} 
\usepackage{wrapfig}

\usepackage[pagebackref,breaklinks,colorlinks]{hyperref}

\usepackage[capitalize]{cleveref}
\crefname{section}{Sec.}{Secs.}
\Crefname{section}{Section}{Sections}
\Crefname{table}{Table}{Tables}
\crefname{table}{Tab.}{Tabs.}

\def\cvprPaperID{*****} 
\def\confName{CVPR}
\def\confYear{2022}

\begin{document}

\title{TextAge: A Curated and Diverse Text Dataset for Age Classification}

\author{Shravan Cheekati\\
{\tt\small
scheekati6@gatech.edu}
\and
 Mridul Gupta\\
{\tt\small mgupta358@gatech.edu}
\and
 Vibha Raghu\\
{\tt\small vraghu6@gatech.edu}
\and
Pranav Raj\\
{\tt\small praj36@gatech.edu}
}

\maketitle


\section{Introduction}
\label{sec:intro}
In recent years, large language models (LLMs) and language-classification models have garnered significant attention due to advancements in Generative Deep Learning and Natural Language Processing abilities. However, these state-of-the-art (SOTA) models often follow a one-size-fits-all approach, generating and classifying text irrespective of the age of the speaker. To close the gap between human-generated and model-generated text, it is crucial to understand the differences in text based on the age of the producer. Moreover, the benefits of classifying text into an age group or producing text based on an age group are widespread. 

Personalized services, such as chatbots, educational tools, healthcare services, and targeted advertisements, will be able to provide information in a way that is best understood by the target age group. The speaking or writing levels of young kids can be evaluated based on how similar their outputs are to that of their peers, or in other words, how likely their produced output belongs to their age group. A similar process can be followed by employers who want to test the language abilities of their employees. Finally, an application we were especially motivated by is Content Moderation. As kids under the age of 13 are rightfully restricted from social media usage, such an ability would be able to detect underage users on social media platforms worldwide. 

We hypothesize that the current lack of this ability to classify text into an age group is due to issues surrounding relevant datasets, rather than insufficient model architectures since SOTA models are able to perform much more complex tasks. For example, to develop an underage detection model, it is crucial to have sufficient data from a younger age group, something that current datasets lack. Hence, we build a comprehensive and diverse dataset, TextAge, that maps a sentence to the age of the producer, age group of the producer, and whether the producer is underage (under 13). As opposed to existing datasets consisting of solely written text, on top of this, our focus was on collecting spoken data from our data sources. In section 2, we elaborate on Related Works. In Section 3, we detail our data collection, cleaning, and labeling process. In Section 4, we apply our dataset to tackle the task of Underage Detection, along with Generation Classification to test the general ability of SOTA models in classifying text by age group.
\section{Related Works}
\label{sec:relatedworks}

Several dataset collections have limitations on their age ranges, motivating us to explore a variety of data sources. We were inspired by the CommonVoice dataset \cite{commonvoice} which consists of speech data from various age groups, but it is limited to individuals over 20 years old and focused on audio classification. The CommonVoice dataset provides sentences for the participants to recite and collects their audio file, hence there is no focus on the context of their words but rather on auditory features. Our dataset is constructed from sources such as CHILDES, Meta, and the TV show "Survivor" to cover a wide range of ages with a focus on context-based speech transcripts and sentences to effectively classify age. 

Additional studies on how age can be determined from text can be studied such as Pennebaker and Stone \cite{pennebaker} who used text samples from writers of different ages to study language changes across the lifespan, finding that older individuals used more positive and future-oriented language. Schwartz et al. \cite{socialmediapersonality} analyzed social media data to investigate how language use varies with age, discovering distinct linguistic patterns for different age groups. Nguyen et al. \cite{twitter} used a large corpus of online forum posts (ex: Twitter/X) to study age-related language variation, identifying key linguistic features that change with age. These studies highlight the growing interest in understanding language patterns across age groups and the need for a comprehensive dataset to support this research. 

Pentel \cite{AutomaticAgeDetect} conducted a similar study in detecting age based on short 100-word texts written by children and adults in Estonian. Logistic Regression, Support Vector Machines, C4.5, k-Nearest
Neighbor, Naïve Bayes, and Adaboost algorithms were compared against each other in performance, however the study was limited to written text rather than spoken transcripts. Blandin et al. \cite{AgeRecommendation} conducts a similar study that classifies the age at which a text is understood by a person, focusing more on reading comprehension rather than any speaking or casual conversation data points. 
Our contributions involve incorporating diverse data sources into a corpus to cover a wider range of ages and language styles to provide a structured dataset for studying age-related language patterns. This is crucial for advancing research in language development, age-related language patterns, and language-based age prediction.

\section{Methods}
\subsection{Data Collection}
\noindent \textbf{CHILDES databank}  \\
To gather conversation data from a younger age group, we selected the CHILDES dataset which focuses specifically on child language development. This dataset was particularly beneficial because of its comprehensive collection of speech data from children. This data is spontaneous and already transcribed. Furthermore, variations of this dataset is well-established in linguistic research studies ensuring reliability of speech data and resting any ethical concerns regarding the use of child data since the transcribed data is completely anonymous.

Initially, we accessed the CHILDES database to retrieve text files containing transcriptions of child speech, focusing on those that met specific criteria related to age and language. Using Python scripts, we then extracted lines specifically attributed to the child speaker, labeled as "CHI" in the transcripts. \newline
 \noindent \textbf{Meta Casual Conversations Dataset}\\
A primary source of a large number of data points originate from the Casual Conversation dataset composing of human annotated transcripts and audio files from over 3000 participants above the age of 18 \cite{metatowards}. Age, gender, apparent skin types, and lighting condition labels were provided by participants. The dataset's intended use is for audio or visual classification as experimented in the DeepFake detection challenge. However, we utilize the transcript data to collect sentences from a wide range of ages. \newline
\noindent \textbf{Poki Poems-by-kids}\\
This dataset consists of written poems written by grade levels 1 to 12 \cite{hipson2020poki}. The lemmatized version of the data was used in order to maintain better normalization of words. Age was determined by taking the lower age range of the input grade. This makes it safer for the impactful use case of online underage detection e.g. a 7th grader would be around 12 to 13 years old so all 7th graders would be assumed to be 12 years old in which case they should not be on social media platforms or other online age restricted platforms. \newline
\noindent \textbf{JUSThink Dialogue and Actions Corpus}\\ 
The JUSThink Dialogue and Actions Corpus (JUSThink) consists of a case study of a robot-mediated human-human collaborative learning activity named JUSThink where youth aged 9 through 12 are recorded and transcripted in their attempt to solve two graph related problems \cite{JustThink}. The specific ages are unable for this dataset so the average floored of 9 to 12 is taken, so the entire dataset of points map to 10 year olds. Since the entire group is under the 13 year old underage cutoff, we determined taking the average would be a better representative instead of flooring the age range to 9. \newline
\noindent \textbf{Survivor}\\
As opposed to scraping from a scripted TV show, we ensured that we added to our dataset from a Reality TV show to collect authentic and real sentences spoken by the contestants and the host. We specifically chose Survivor as we found organized transcripts for each episode for the first 40 seasons of the show, resulting in around 160000 entries. For the age label, we manually found the age of each of the ~20 contestants per season.
\subsection{Data Cleaning}
The data cleaning process was important in preparing the respective data sources for analysis, focusing on enhancing data quality and consistency. Initially, transcripts were retrieved as raw text and required significant cleaning to remove transcription errors, non-verbal cues, and other irrelevant metadata. To address this we developed cleaning scripts using Python, utilizing the Pandas and regex library. We used regular expressions to strip out non-alphabetic characters, correct punctuation misplacements, remove placeholder text, remove time stamps, and other annotation symbols. Additionally, any sentences reduced to single words or non-meaningful fragments were discarded to maintain analytical relevance. The cleaned data was then structured into two primary columns, 'Sentence' and 'Age', and merged from multiple sources into a single CSV file, ensuring a uniform dataset for subsequent analysis.

\subsection{Data Exploration}

\begin{figure}[htbp] 
  \centering
  \includegraphics[width=0.6\linewidth]{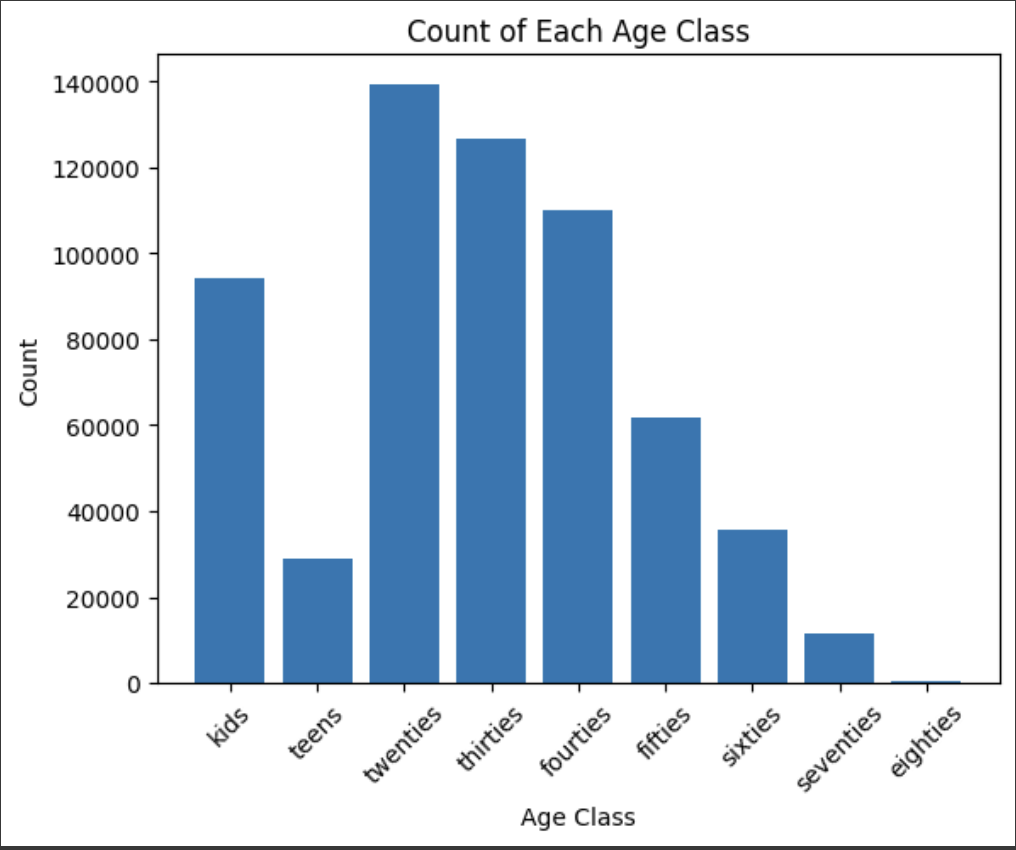} 
  \caption{Distribution by Age Group of our Final Dataset. Highlights less data for teens and older age groups.} 
  \label{fig:ageDist} 
\end{figure}

\begin{figure}[htbp] 
  \centering
  \includegraphics[width=0.6\linewidth]{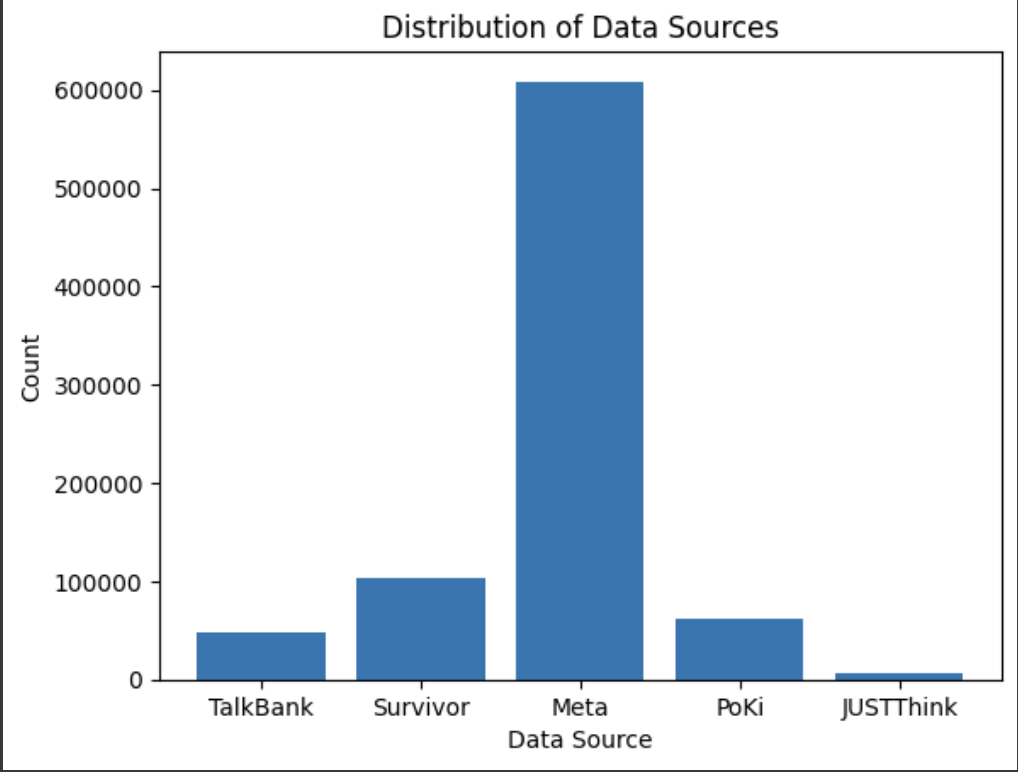} 
  \caption{Visualizes how each source contributes to our final dataset of 608082 rows. Count represents number of entries per source before cleaning.} 
  \label{fig:dataDist} 
\end{figure}

On top of the two primary columns, we added two additional columns to our dataset: underage (True or False) and age-group (kids, teens, twenties, thirties, fourties, fifties, sixties, seventies, eighties). The motivation behind this was to prepare our dataset for potential classification tasks. We then performed an extensive Data Exploration process, where we analyzed surface-level differences (average length of sentences, number of unique words, etc.) between age groups. We also created Word Clouds to help in a visual understanding of our massive dataset. Figure \ref{fig:ageDist} and \ref{fig:dataDist} below are key results from our Data Exploration.

\section{Applications}

\begin{figure*}[!htb]
    \centering
    \begin{subfigure}{0.3\textwidth}
        \centering
        \includegraphics[width=\textwidth]{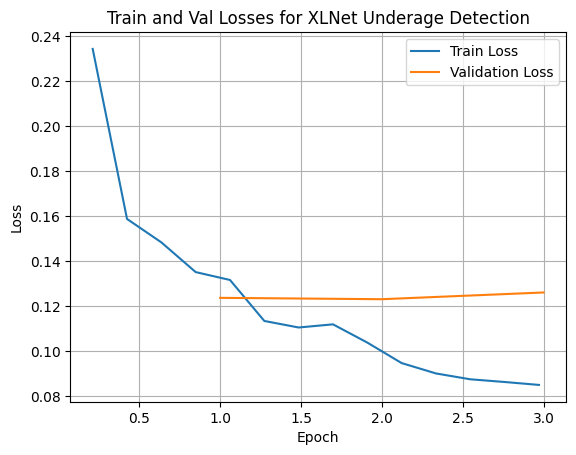}
        \caption{XLNet train loss (binary)}
    \end{subfigure}
    \hfill
    \begin{subfigure}{0.3\textwidth}
        \centering
        \includegraphics[width=\textwidth]{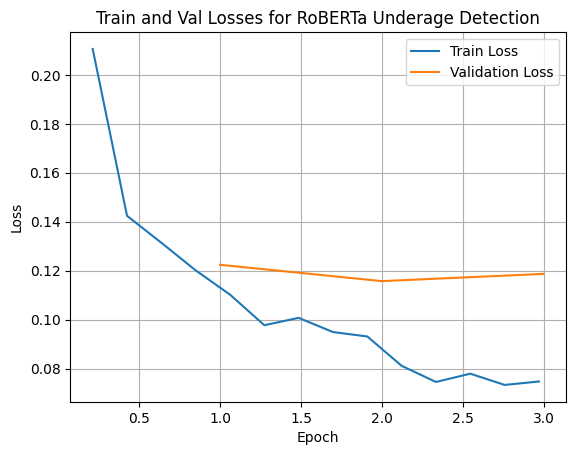}
        \caption{RoBERTa train loss (binary)}
    \end{subfigure}
    \hfill
    \begin{subfigure}{0.3\textwidth}
        \centering
        \includegraphics[width=\textwidth]{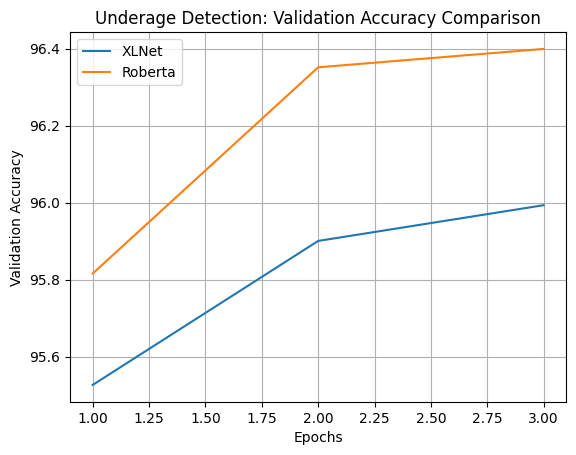}
        \caption{XLNet vs RoBERTa val acc. (binary)}
    \end{subfigure}
    
    \vspace{1em}
    
    \begin{subfigure}{0.3\textwidth}
        \centering
        \includegraphics[width=\textwidth]{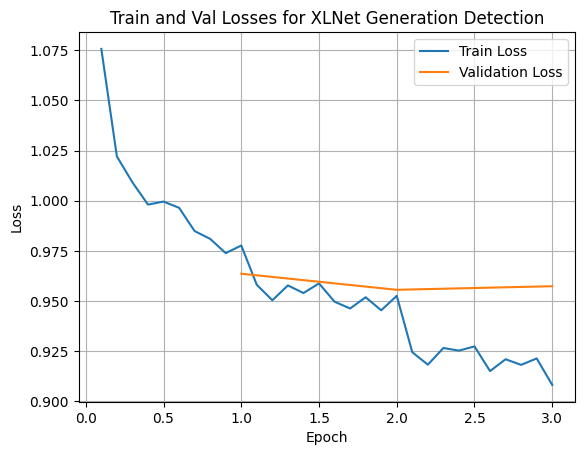}
        \caption{XLNet train loss (multiclass)}
    \end{subfigure}
    \hfill
    \begin{subfigure}{0.3\textwidth}
        \centering
        \includegraphics[width=\textwidth]{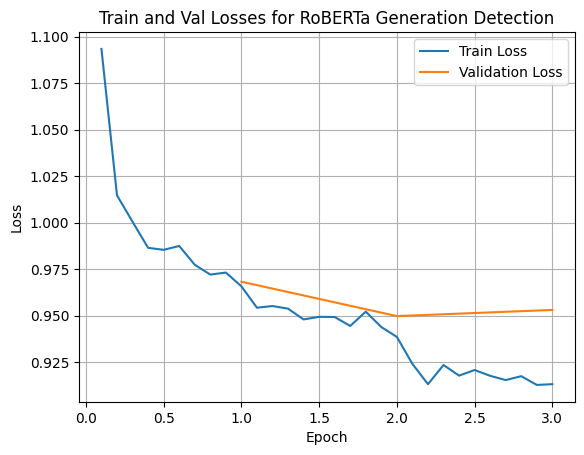}
        \caption{RoBERTa train loss (multiclass)}
    \end{subfigure}
    \hfill
    \begin{subfigure}{0.3\textwidth}
        \centering
        \includegraphics[width=\textwidth]{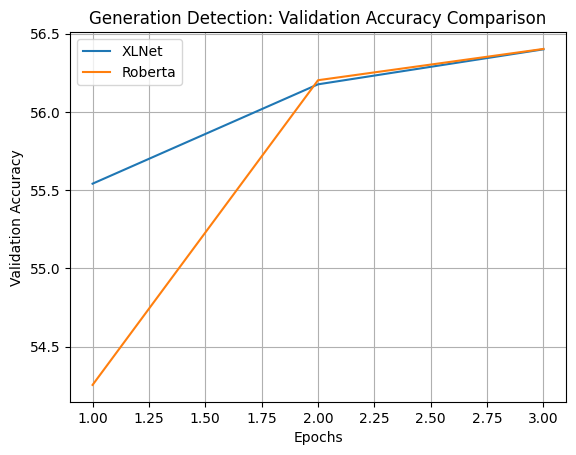}
        \caption{XLNet vs RoBERTa val acc. (multiclass)}
    \end{subfigure}
    
    \caption{XLNet and RoBERTa models performance metrics on binary and multiclass tasks.}
    \label{fig:results}
\end{figure*}

We present two applications of our age-diverse language dataset: Underage Detection and Generational Classification. These tasks demonstrate the utility of our dataset in identifying age-related linguistic patterns and showcase its potential for various real-world applications.

Underage Detection is a binary classification task that aims to differentiate between language patterns characteristic of minors (individuals under the age of 13) and young-adults and over (individuals aged 13 and above). This task has important implications in online safety, content moderation, and age-appropriate communication. By accurately identifying the age group of a speaker or writer based on their language use, online platforms can better protect minors from inappropriate content and interactions. 

Generational Classification is a multiclass classification task that seeks to categorize language patterns into different generational age groups, such as kids, teens, twenties, thirties, and so on. Understanding the linguistic differences between age groups can provide valuable insights into language development, social trends, and age-related preferences. This information can be applied in various domains, including targeted marketing, product design, and social research. 
\begin{table}[htbp]
\centering
\begin{tabular}{|l|l|l|}
\hline
\textbf{Model} & \textbf{Binary} & \textbf{Multiclass} \\ \hline
Naive Bayes & 0.8798 & 0.3505 \\ \hline
Roberta & 0.9640 & 0.5462 \\ \hline
XLNet & 0.9599 & 0.5422 \\ \hline
\end{tabular}
\caption{F1 Scores for both underage and generation detection}
\label{tab:overall_f1_score}
\end{table}
\subsection{Underage/Of-Age Detection}
We trained three models for the Underage Detection task: a Naive Bayes classifier as a baseline, and fine-tuned RoBERTa and XLNet models. The Naive Bayes classifier served as a simple, probabilistic baseline for comparison. RoBERTa and XLNet, both state-of-the-art transformer-based models, were fine-tuned on our dataset to capture more complex linguistic patterns associated with age groups.
\subsubsection{Results \& Analysis}
The fine-tuned RoBERTa model achieved the highest f1 score of 0.9640 and test accuracy of 0.9640, closely followed by the fine-tuned XLNet model with an f1 score of 0.9599 and test accuracy of 0.9599. Both transformer-based models significantly outperformed the Naive Bayes baseline, which obtained an f1 score of 0.8798 and test accuracy of 0.8799.\\
Figure \ref{fig:results} displays the training loss curves for the RoBERTa and XLNet models over the fine-tuning epochs. Both models exhibit a steady decrease in training loss, indicating effective learning and adaptation to the age-related linguistic patterns present in our dataset.

The results demonstrate the power of the fine-tuned RoBERTa and XLNet models over the Naive Bayes baseline for the Underage Detection task. The transformer-based models' ability to capture complex linguistic patterns and contextual information contributed to their higher performance. The fine-tuning process allowed these models to adapt their pre-trained language understanding to the specific age-related patterns in our dataset.
The strong performance of RoBERTa and XLNet highlights the potential of using advanced language models for accurate age group classification based on linguistic cues. This has significant implications for enhancing online safety measures, content moderation, and age-appropriate recommendations in various digital platforms.
\subsection{Generation Classification}
For the Generational Classification task, we employed the same three models as in the Underage Detection task. The methodology remained similar, with the models being trained to classify language patterns into different age groups.
\subsubsection{Results \& Analysis}
The classification reports for the Naive Bayes, RoBERTa, and XLNet models are presented in the appendix [\ref{tab:multiclass_f1_score}]. The RoBERTa model achieved the highest overall f1 score of 0.5462, closely followed by the XLNet model with an f1 score of 0.5422. The Naive Bayes baseline performed significantly worse, with an f1 score of 0.3505.

Looking at the individual age groups, all three models performed exceptionally well in classifying the "kids" group, with f1 scores above 0.9 for RoBERTa and XLNet. The performance on the "teens" and "twenties" groups was moderate, with f1 scores ranging from 0.19 to 0.58. For age groups above 30, the models' performance varied, with the Naive Bayes classifier struggling across all older age groups, while RoBERTa and XLNet showed better performance for the "thirties" and "fourties" groups.
However, the models struggled to classify the "fifties," "sixties," and "seventies" age groups effectively. RoBERTa and XLNet achieved low f1 scores for these groups, and the Naive Bayes classifier performed poorly as well. It is worth noting that the support for these older age groups was relatively lower compared to the younger age groups and so there may have been a significant shortage of training data for these groups.

The Generational Classification task proved to be more challenging than the Underage Detection task, with lower f1 scores and accuracies achieved by the models, suggesting that capturing fine-grained linguistic differences between age groups is more complex. While the models performed strongly on the "kids" group, indicating distinctive language patterns, performance decreased for older age groups, particularly "fifties," "sixties," and "seventies," likely due to limited data samples and less pronounced linguistic differences. To improve performance, future work could focus on collecting more diverse data for older age groups and exploring advanced modeling techniques. Despite the challenges, the fine-tuned RoBERTa and XLNet models demonstrated reasonable performance, outperforming the Naive Bayes baseline and highlighting the potential of advanced language models for granular age group classification.
\section{Conclusion}
As touched upon in the applications section, the use-cases of the dataset we curated are massive. In addition to classification tasks, our dataset can be used in Seq2Seq tasks where a given sentence is converted to text belonging to a certain age group. This would be specifically useful for targeted advertisements. While our current dataset is ready to use for a variety applications (reach out through email), we plan to expand our dataset to create an even more diverse and comprehensive corpus.
{\small
\bibliographystyle{ieee_fullname}
\bibliography{egbib}
}

\section{Appendix}

\begin{table}[htbp]
\centering

\begin{tabular}{|l|l|l|l|l|}
\hline
\textbf{Model} & \textbf{Class} & \textbf{Precision} & \textbf{Recall} & \textbf{F1-Score} \\ \hline
\multirow{2}{*}{Naive Bayes} & Underage & 0.87 & 0.90 & 0.88 \\ \cline{2-5} 
 & Of Age & 0.90 & 0.86 & 0.88 \\ \hline
\multirow{2}{*}{Roberta} & Underage & 0.97 & 0.96 & 0.96 \\ \cline{2-5} 
 & Of Age & 0.96 & 0.97 & 0.96 \\ \hline
\multirow{2}{*}{XLNet} & Underage & 0.97 & 0.95 & 0.96 \\ \cline{2-5} 
 & Of Age & 0.95 & 0.97 & 0.96 \\ \hline
\end{tabular}
\caption{Underage detection f1 score breakdown}
\label{tab:f1_score}
\end{table}

\begin{table}[htbp]
\centering

\begin{tabular}{|l|l|l|l|l|}
\hline
\textbf{Model} & \textbf{Class} & \textbf{Precision} & \textbf{Recall} & \textbf{F1-Score} \\ \hline
\multirow{9}{*}{Naive Bayes} & Kids & 0.05 & 0.01 & 0.02 \\ \cline{2-5} 
 & Teens & 0.28 & 0.06 & 0.09 \\ \cline{2-5} 
 & Twenties & 0.36 & 0.29 & 0.32 \\ \cline{2-5} 
 & Thirties & 0.66 & 0.78 & 0.71 \\ \cline{2-5} 
 & Fourties & 0.15 & 0.01 & 0.01 \\ \cline{2-5} 
 & Fifties & 0.27 & 0.04 & 0.07 \\ \cline{2-5} 
 & Sixties & 0.35 & 0.17 & 0.23 \\ \cline{2-5} 
 & Seventies & 0.29 & 0.29 & 0.29 \\ \cline{2-5} 
 & Eighties & 0.33 & 0.59 & 0.43 \\ \hline
\multirow{8}{*}{Roberta} & Kids & 0.89 & 0.98 & 0.93 \\ \cline{2-5} 
 & Teens & 0.66 & 0.11 & 0.19 \\ \cline{2-5} 
 & Twenties & 0.51 & 0.65 & 0.57 \\ \cline{2-5} 
 & Thirties & 0.49 & 0.39 & 0.44 \\ \cline{2-5} 
 & Fourties & 0.56 & 0.61 & 0.58 \\ \cline{2-5} 
 & Fifties & 0.50 & 0.28 & 0.36 \\ \cline{2-5} 
 & Sixties & 0.00 & 0.00 & 0.00 \\ \cline{2-5} 
 & Seventies & 0.00 & 0.00 & 0.00 \\ \hline
\multirow{8}{*}{XLNet} & Kids & 0.88 & 0.97 & 0.92 \\ \cline{2-5} 
 & Teens & 0.64 & 0.12 & 0.19 \\ \cline{2-5} 
 & Twenties & 0.50 & 0.69 & 0.58 \\ \cline{2-5} 
 & Thirties & 0.49 & 0.35 & 0.41 \\ \cline{2-5} 
 & Fourties & 0.56 & 0.61 & 0.59 \\ \cline{2-5} 
 & Fifties & 0.54 & 0.27 & 0.36 \\ \cline{2-5} 
 & Sixties & 0.00 & 0.00 & 0.00 \\ \cline{2-5} 
 & Seventies & 0.00 & 0.00 & 0.00 \\ \hline
\end{tabular}
\caption{Generation detection F1 Score breakdown}
\label{tab:multiclass_f1_score}
\end{table}

\end{document}